\documentclass{article}
\PassOptionsToPackage{round}{natbib}
\usepackage[dblblindworkshop, final]{neurips_2026}
\workshoptitle{Long-Context Foundation Models}
\usepackage[utf8]{inputenc}
\usepackage[T1]{fontenc}
\usepackage{amsmath,amssymb,amsthm}
\usepackage{amsfonts}
\usepackage{booktabs}
\usepackage{graphicx}
\usepackage{microtype}
\usepackage{bm}
\usepackage[hidelinks]{hyperref}
\usepackage{url}
\usepackage{xcolor}

\newcommand{\R}{\mathbb{R}}
\newcommand{\half}{\tfrac12}
\newcommand{\detJ}{\det J}

\newcommand{\aff}[1]{\textsuperscript{\normalfont #1}}

\usepackage{fancyhdr}
\fancypagestyle{workshophead}{%
  \fancyhf{}%
  \fancyhead[C]{\footnotesize 40th Conference on Neural Information Processing Systems (NeurIPS 2026). Workshop: Long-Context Foundation Models.}%
  \fancyfoot[C]{\thepage}%
  }
\title{Certified Access: Test-Time Compute on Conservative Physics-Structured AI Memory}
\author{%
  Pratik Jawahar \aff{1,2}\thanks{Corresponding author: \texttt{pratik.jawahar@cern.ch}}
  \And Riccardo Maggioni \aff{3}
  \And Jonas Petersen \aff{3}
  \And Matteo Cozzi \aff{3}
  \And Gian-Alessandro Lombardi \aff{2}
  \And Marek Gajewski \aff{2}
  \And Camilla Mazzoleni \aff{3}
  \And Federico Martelli \aff{2}
  \AND
  \normalfont \aff{1} CERN \quad \aff{2} Forgis AG \quad \aff{3} ETH Zurich
}
\date{}

\begin{document}
\sloppy
\maketitle
\thispagestyle{workshophead}

\begin{abstract}
Test-time compute mechanisms, such as retries, confidence-based escalation and non-local routing, are conventionally deployed as black-box additions to trained models. They typically add forward passes or halting heads, but give no structural account of what the extra compute buys or what it costs against the model's stability guarantees. We therefore propose a theoretical and empirical framework for test-time compute on a Latent Dynamics Model used as the memory. When an encoder's latent state $(q,p)$ is advanced by a damped symplectic (velocity-Verlet) step of a learned Hamiltonian, every mechanism that expands access has an exact phase-volume Jacobian, a bounded energy change and an exact latch-transport law. These certificates are conditional: the volume certificate holds for damping frozen within a step, not for the state-dependent governor, and the energy statements hold on the stated testbed potentials. We split reachability failures into escape failures, where a target inside a kinematic causal box $C_T$ is trapped behind a latent barrier, and reach failures, where the target lies outside $C_T$. A Lorentz squeeze resolves escape failures by injecting bounded, re-absorbable energy. A gated canonical translation (a one-step channel across $C_T$) resolves reach failures at $\detJ=1$ exactly. We verify this certificate stack on a designed analytic testbed. Reaching just beyond $C_T$ with the squeeze requires energy that grows quadratically in the excess distance. The gated channel's energy cost equals the potential difference between its endpoints, which is zero on the matched channel measured here, and so does not grow with distance in this testbed. An untrained state-replacing map annihilates phase volume and erases stored charge, while the gated channel transports it. On a learned physics-structured memory, we show calibrated compute rationing and evaluate the scaling limits of non-local routing against a physics-free baseline. The main contribution is the formalization and verification of the certificate stack, not benchmark performance.
\end{abstract}

\section{Introduction}
Deep learning models increasingly allocate inference compute adaptively. Learned halting (ACT, \citealp{graves_adaptive_2017}; PonderNet, \citealp{banino_pondernet_2021}), confidence-gated early exits (CALM, \citealp{schuster_confident_2022}), learned routing (Mixture-of-Depths, \citealp{raposo_mixture--depths_2024}; Mixture-of-Experts, \citealp{shazeer_outrageously_2017}) and energy-based verification \citep{gladstone_energy-based_2025} have established the value of input-conditional escalation. However, these mechanisms generally attach a learned scalar gate to a feedforward stack. Escalation then repeats the standard forward operation and gives no structural guarantee of how the extra computation changes the model's latent state.

We therefore define test-time compute for Latent Dynamics Models (LDMs), the class of models that structure the dynamics of an abstract latent space produced by an encoder. Suppose the queried memory is a conservative latent state $(q,p)$, produced by the encoder and advanced by a structure-preserving symplectic integrator of a learned separable Hamiltonian $H(q,p)=T(p)+V_\theta(q)$. This Hamiltonian describes the latent space of any encoder and is not limited to the input system. Test-time compute then becomes access to a defined phase space at inference. As in~\citet{jawahar_chlu_2026}, the kinetic term is a causal relativistic governor. In this regime the properties of the phase space define every access mechanism. We call this certified access: each mechanism carries a physical certificate.

The certificate has three parts: the Jacobian determinant $\detJ$, which certifies phase-volume conservation; a bounded energy change, which quantifies injected energy; and a latch-transport law, which states how content stored in protected continuous-symmetry directions transforms. These certificates are theorems that follow from the symplectic structure under stated assumptions (damping fixed within a step, and the testbed potentials). They are not empirical curves fit to data, and they are not guarantees for the full learned system. We first state each certificate and verify it on a designed testbed (\S\ref{sec:mech}): the exactness claims ($\detJ$, the energy change and the latch-transport law) hold to machine precision, while the re-absorption and channel-hop thresholds hold to leading order. We then place the mechanisms on learned memories (\S\ref{sec:learned}).

This work makes six contributions:
\begin{enumerate}\itemsep2pt
\item Formalization of failure modes. We partition reachability failures into two kinds. In kinematic reach failure, the target lies outside a causal box $C_T$ bounded by $L_i=T\varepsilon c/\sqrt{M_i}$. In energetic escape failure, the target is inside $C_T$ but shielded by a potential barrier. Distinct mechanisms, with different certificates, resolve the two (\S\ref{sec:setup}).
\item Verification of the certificate stack. We verify our theoretical predictions on an analytic testbed with dimensions 2 and 4, 5 seeds, and oracle placement. The Lorentz squeeze cures escape with bounded injection and extends reach at an energy exponential in the rapidity. The gated channel cures reach continuously with $\detJ=1$, exact latch transport, and an energy change equal to the endpoint potential difference, zero on the matched channel measured here (\S\ref{sec:mech}).
\item Empirical measurement of latch transport. We show that an untrained state-replacing map has $\detJ=0$ exactly and erases the stored Goldstone charge. The certified gated channel instead transports the charge with preserved spread. Phase volume alone does not suffice for this transport; a matched channel is required. A bounded energy change is also necessary but not sufficient for Bounded-Input Bounded-Output (BIBO) stability without coercive-component screening (\S\ref{sec:erasure}, \S\ref{sec:bibo}).
\item Calibrated compute rationing. We implement a self-calibrating compute-rationing gate on an escalatable learned memory. The gate reduces relaxation steps by $4.81\pm0.44\times$ at capacity kv16 with a 400-epoch budget. Its value lies in escalatability and rationing, not in a better baseline energy signal (\S\ref{sec:rationing}).
\item Scaling limits of non-local routing. We verify that a direct one-hop non-local edge keeps a flat FLOP count as distance grows, unlike multi-hop diffusion. We also report that a parameter-matched, physics-free router outperforms the energy-gated version of this mechanism in both floating-point operations (FLOPs) and accuracy (\S\ref{sec:routing}).
\item Capacity and epoch-budget map of the compute dial. We map the rationing gate across capacity and training budget. The measured step reductions ($9.9\times, 9.5\times, 6.2\times$ across kv32, kv64, kv96) are intra-CLU rationing against a full-budget CLU baseline. Beyond kv64, the epoch budget, not a hard capacity limit, constrains performance. Gate accuracy degrades sharply under evaluation cue noise, while storage fidelity stays at $1.0$ (\S\ref{sec:regime}).
\end{enumerate}

\section{Setup: A Conservative Memory and its Reachable Set}
\label{sec:setup}
Our reference memory is the Causal Learning Unit (CLU), introduced as CHLU in \citet{jawahar_chlu_2026}. A CLU is a latent dynamics model whose state is a phase-space point $(q,p)$: an encoder writes the state, and a learned Hamiltonian evolves it with a damped symplectic integrator, so retrieval is relaxation under learned dynamics. Our empirical results use the CLU's reference training objective but measure general properties of the learned Hamiltonian.

One CLU step, the governed map, advances $(q,p)$ by a dissipative velocity-Verlet step~\citep{martys1999velocity} with step size $\varepsilon$ and per-step damping $\gamma\in[0,1)$. Velocity-Verlet is a kick--drift--kick scheme: a half-step momentum update (kick), a full position update (drift), then a second half-step kick. Here it integrates the separable Hamiltonian $H=T(p)+V_\theta(q)$ and is followed by the momentum scaling $p\mapsto(1-\gamma)p$. For fixed $\gamma$ this update is conformally symplectic, $J^\top\Omega J=(1-\gamma)\Omega$, so $\detJ=(1-\gamma)^d$ in $d$ dimensions. In relativistic mode the reference architecture's kinetic energy is
\[
T(p)=\sqrt{c^2\,p^\top M^{-1}p+m_0^2c^4},
\]
with $M=\mathrm{diag}(M_1,\dots,M_d)\succ0$ the learned inertial mass, $m_0>0$ the rest mass (values in the flag table, App.~\ref{app:flags}) and $c$ a hyperparameter. This kinetic mode caps velocity at $|\dot q_i|<c/\sqrt{M_i}$. To bleed excess energy toward a target $E^\star$, a state-dependent governor sets $\gamma_n=s\tanh(\max(0,H-E^\star))$. The determinant certificate holds for the damping value frozen within a step. Because $\gamma_n$ depends on the state, its derivative adds terms to the Jacobian, and the certificate does not cover the governed map.

The latent reachable set of the governed map $\Phi_{\varepsilon,\gamma}$ over $T$ steps from $z_0$ is $R_T(z_0)=\{\Phi^n(z_0):1\le n\le T\}$, with position shadow $Q_T(z_0)$. A mechanism gives access to a latent state only if it enlarges $Q_T$. Throughout, $z=(q,p)\in\R^d\times\R^d$.

The causal box follows from a bound on the drift: one drift sub-step advances each coordinate by strictly less than $\varepsilon c/\sqrt{M_i}$, for every momentum $p$ and every potential $V_\theta$. The drift sub-step of the kick--drift--kick map moves only the position, $q\mapsto q+\varepsilon\,\nabla_pT(p)$, at whatever momentum is current at that sub-step. Differentiating $T$,
\[
\frac{\partial T}{\partial p_i}=\frac{c^2p_i}{M_i\,T(p)},
\qquad
T(p)=\Bigl(c^2\sum\nolimits_j p_j^2/M_j+m_0^2c^4\Bigr)^{1/2}>\frac{c\,|p_i|}{\sqrt{M_i}},
\]
the last inequality strict for every $p$ because $m_0>0$ and the remaining kinetic terms are non-negative. Hence
\[
|\dot q_i|=\Bigl|\frac{\partial T}{\partial p_i}\Bigr|=\frac{c^2|p_i|}{M_i\,T(p)}<\frac{c^2|p_i|}{M_i}\cdot\frac{\sqrt{M_i}}{c\,|p_i|}=\frac{c}{\sqrt{M_i}},
\]
so a single drift obeys $|q_i'-q_i|<\varepsilon c/\sqrt{M_i}$. The two kick sub-steps and the damping $p\mapsto(1-\gamma)p$ act on momentum only and move no position. The same bound therefore holds for the full step, uniformly in $\theta$, $\gamma$, and the state. Summing over $n\le T$ steps gives $|q_{n,i}-q_{0,i}|<n\varepsilon c/\sqrt{M_i}\le L_i$, so the reachable set is bounded:
\[
Q_T(z_0)\ \subseteq\ C_T(q_0):=\{q:\ |q_i-q_{0,i}|\le L_i,\ L_i=T\varepsilon c/\sqrt{M_i}\}.
\]

The bound is energy-blind: it involves only $(\varepsilon,c,M_i)$, and neither $V_\theta$ nor the magnitude of $p$ enters. Injected momentum drives $|\dot q_i|$ toward the cap $c/\sqrt{M_i}$ from below but never past it. The relative deficit is $1-|\dot q_i|\sqrt{M_i}/c\approx m_0^2c^2M_i/(2p_i^2)$ as the momentum $|p_i|$ along coordinate $i$ grows, so the velocity saturates at the cap. In Newtonian mode $\nabla_pT=M^{-1}p$ is unbounded, no such box exists, and reach is energy-limited. This is why the Newtonian control in \S\ref{sec:testbed} crosses the box.

This setup has two failure modes. A \emph{kinematic reach} failure occurs when the target $q^\star\notin C_T(q_0)$; only a non-local jump solves it. An \emph{energetic escape} failure occurs when $q^\star\in C_T(q_0)$ but is shielded by a potential barrier $\Delta V_b$ that the latent energy cannot cross. Any bounded energy injection that acts faster than the governor dissipates it resolves this failure.

\section{Mechanism Certificates and Limitations}
\label{sec:mech}
We first certify the access mechanisms and the latch, then state where the certificate stops (BIBO stability), and finally compare the mechanisms on a testbed.

To resolve escape failures, we apply a mass-weighted Lorentz squeeze $S^{(M)}_\zeta$, which injects energy to cross barriers in the spirit of basin-hopping \citep{wales_global_1997}. On the targeted coordinate pair $(q_i,p_i)$, in mass-weighted coordinates $u=\sqrt{M_i}\,q_i$, $v=p_i/\sqrt{M_i}$, the squeeze acts as the hyperbolic rotation $u'=u\cosh\zeta+v\sinh\zeta$, $v'=v\cosh\zeta+u\sinh\zeta$, and as the identity on all other coordinates. The map is linear with determinant $\cosh^2\zeta-\sinh^2\zeta=1$, hence exactly symplectic ($\det=1$). It amplifies velocity and displacement toward the saddle, $\delta q'=\delta q\cosh\zeta+p_0\sinh\zeta$, where the rapidity $\zeta$ sets the squeeze strength. For a Hamiltonian that is quadratic and isotropic in the same mass-weighted coordinates, $H_{\mathrm q}=\half\kappa\,(u^2+v^2)$, the key step is
\[
u'^2+v'^2=(u^2+v^2)\cosh2\zeta+2uv\,\sinh2\zeta
\ \le\ (u^2+v^2)\,e^{2|\zeta|},
\]
using $2|uv|\le u^2+v^2$. The injected energy therefore obeys $H(S_\zeta z)\le e^{2|\zeta|}H(z)$ (full derivation in App.~\ref{app:deriv:squeeze}), and the governor dissipates the excess in about $2\zeta/\gamma_c$ steps. This bound is proved for a Hamiltonian quadratic in position and momentum, whereas the causal box uses the relativistic kinetic energy, so the two results rest on different kinetic energies and are not a single joint statement.

To resolve reach failures, we introduce an intra-unit \emph{gated channel}: a learned pair of loci $(a,b)$ with a hard gate frozen at capture. The channel applies the constant canonical translation $q\mapsto q+\Delta$, $p\mapsto p$, with $\Delta=b-a$. This translation is the flow of the linear Hamiltonian $W=\Delta\cdot p/\tau$, so the Jacobian is $I_{2d}$ and $\detJ=1$ exactly, independent of $\Delta$ and of the gate. The energy cost is the discrete jump $\Delta H_{\rm wh}=V_\theta(q+\Delta)-V_\theta(q)$, the potential difference between the endpoints, which must be tracked; matched loci give zero-energy transport.

We next derive the Jacobian when the gate may vary during the jump, $q\mapsto q+g(q)\Delta$, $p\mapsto p$. This Jacobian is block-triangular with momentum block $I_d$ and position block the rank-one update $\partial q'/\partial q=I_d+\Delta\,\nabla g(q)^\top$. By the matrix determinant lemma,
\[
\detJ=\det\bigl(I_d+\Delta\,\nabla g(q)^\top\bigr)=1+\nabla g(q)^\top\Delta\neq1 .
\]
The defect is first order in the gate's variation along the jump, so no smallness of $\Delta$ removes it (unit-test value $2.05$). Freezing the gate at capture ($g$ constant during the jump) restores $\detJ=1$ exactly, so the gate must stay frozen. This is a per-branch statement (gate fixed); it does not show that the full routing policy is invertible.

The latch transport law concerns a protected direction with the invariant $Q=p^\top X q$, where $X$ is the generator of the broken symmetry. We call $Q$ the Goldstone charge, borrowing from field theory along the lines of~\citet{iqbal_spontaneous_2026}. The channel shifts it by exactly $Q'-Q=p^\top X\Delta$, so the charge is preserved if and only if $\Delta\perp X^\top p$. Because the channel maps a single phase point, it transports the latch rather than copying or erasing it. This law describes the jump only: the damping substep then scales $p$, and with it $Q$, so the charge is not preserved over a full damped retrieval.

Table~\ref{tab:cert} collects the certificate per mechanism. Both jump mechanisms must be paired with a coercive potential $V_\theta$. Deep convolutional potentials are non-coercive out-of-unit by architecture, so exits must lie inside a coercive sub-level component (\S\ref{sec:bibo}).

\begin{table}[htb!]\centering\footnotesize
\caption{Certificate per mechanism. The $(1-\gamma)^d$ entry holds for the damping value frozen within a step; the state-dependent governor is not covered by it.}
\label{tab:cert}
\begin{tabular}{p{2.5cm}p{2.2cm}p{1.8cm}p{2.9cm}p{2.3cm}}
\toprule
Mechanism & Energy Injected & Volume $\detJ$ & Governor Re-absorption & Latch Impact \\
\midrule
Gated channel (translation) & $\Delta V{=}V(b){-}V(a)$, discrete jump & 1 exactly & $\approx(1/\gamma_c)\ln(1{+}\Delta V/E^\star)$ & Transport via shift $p^\top X\Delta$ \\
Gated channel (throat) & standing wave, no jump & $(1{-}\gamma)^d$ (fixed $\gamma$) & continuous & adds $\mu^2{\propto}A/\ell^2$ \\
Squeeze $S^{(M)}$ & $\le(e^{2|\zeta|}{-}1)H$ & 1 exactly & $\approx2\zeta/\gamma_c$ & preserved \\
\bottomrule
\end{tabular}
\end{table}

\subsection{BIBO Stability Limits}
\label{sec:bibo}
\begin{figure}[htb!]\centering
\includegraphics[width=0.6\linewidth]{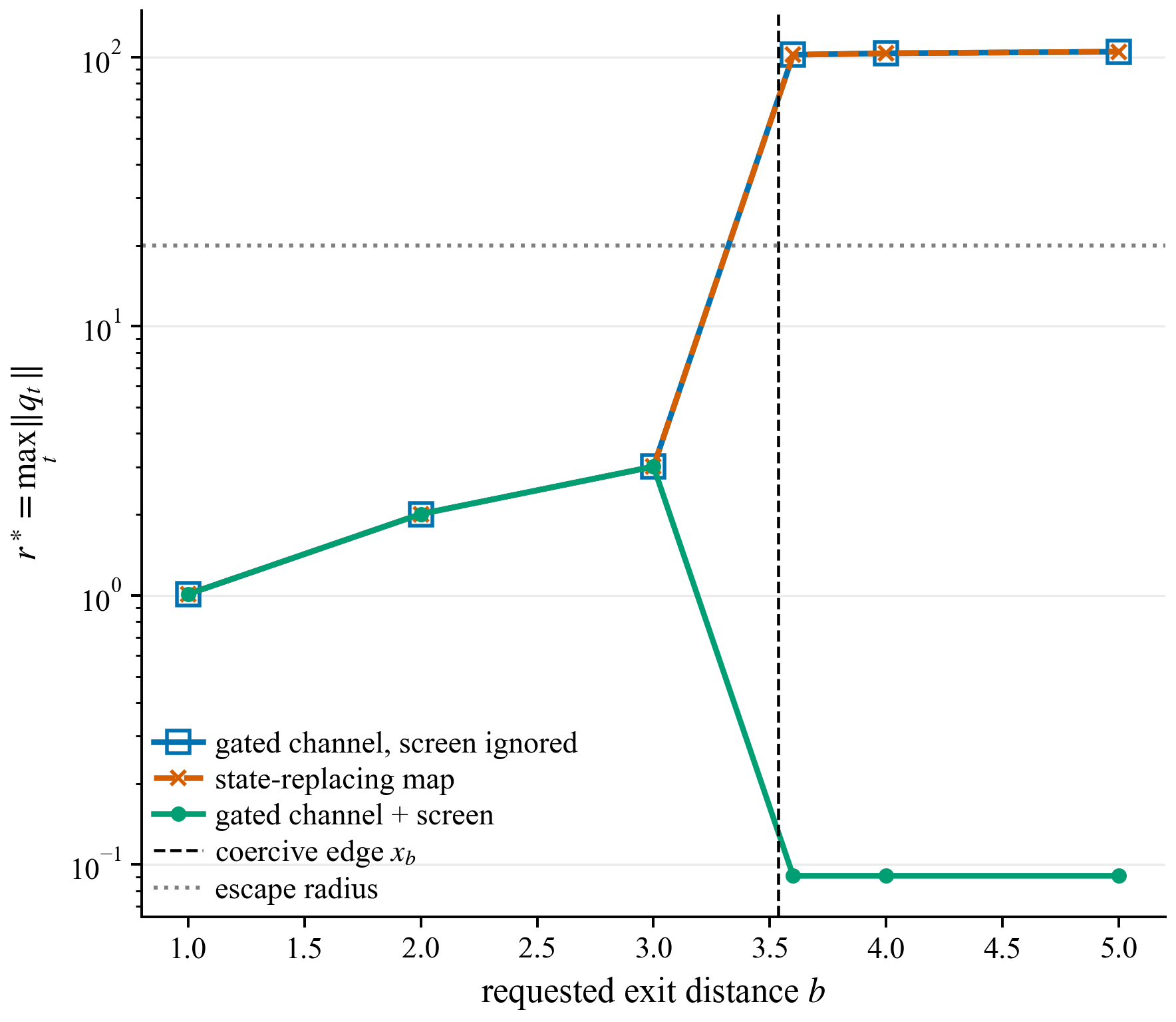}
\caption{The BIBO battery. Maximum excursion radius
$r^\ast=\max_t\|q_t\|$ at horizon $2T$, against the requested exit locus $b$. Exits inside the
coercive component stay bounded for every arm. Beyond the coercive edge $x_b=3.54$ the
unscreened arms grow as $r^\ast\propto T$, while the screened gated channel holds at $r^\ast=0.09$
because it \emph{refuses} the jump rather than making it safe. The state-replacing map
coincides exactly with the screen-ignored ablation, so the screen, not the jump, provides
boundedness. Note $b=5.0$: the energy change is exactly zero and the exit still escapes.}
\label{fig:bibo}
\end{figure}

The certificate $\detJ=1$ guarantees volume preservation, not bounded trajectories: an exit outside the coercive connected component of $V_\theta$ can escape to infinity even when the jump is symplectic and costs zero energy. We test boundedness on a controlled non-coercive well, $V=\half kq_0^2-\epsilon q_0^4$, whose connected component spans $|q_0|<x_b=3.536$. We measure the maximum excursion radius $r^\ast$ at horizons $T$ and $2T$ (Table~\ref{tab:bibo}, Fig.~\ref{fig:bibo}).

\begin{table}[htb!]\centering\small
\caption{The BIBO battery: maximum excursion radius $r^\ast$ at $2T$ against the exit locus $b$, escape rates, and the admit/reject decision of each screen.}
\label{tab:bibo}
\begin{tabular}{lcccccc}
\toprule
Algorithm ($r^\ast$ at $2T$) & $b{=}1.0$ & $2.0$ & $3.0$ & $3.6$ & $4.0$ & $5.0$ \\
\midrule
Gated channel with certificate & $1.01$ & $2.01$ & $3.01$ & $0.09$ & $0.09$ & $0.09$ \\
Gated channel, ablation & $1.01$ & $2.01$ & $3.01$ & $102.13$ & $103.43$ & $104.83$ \\
State-replacing map & $1.01$ & $2.01$ & $3.01$ & $102.13$ & $103.43$ & $104.83$ \\
\midrule
Escape rate, certified & $0$ & $0$ & $0$ & $0$ & $0$ & $0$ \\
Escape rate, blind & $0$ & $0$ & $0$ & $1.0$ & $1.0$ & $1.0$ \\
Gated channel certificate logic & ADMIT & ADMIT & ADMIT & REJECT & REJECT & REJECT \\
Energy sub-level test & admit & admit & admit & reject & admit & admit \\
\bottomrule
\end{tabular}
\end{table}

An admissible exit at $b=3.0$ costs $\Delta H=2.88$ but stays bounded. The escaping arms grow linearly, with $r^\ast(2T)/r^\ast(T)\approx1.9$. These trajectories are unbounded and limited only by the terminal velocity cap. The key case is exit $b=5.0$. There the energy change is exactly zero ($\Delta H=0.0$), so an energy-only sub-level test wrongly admits the jump, and the state escapes at once. This shows that $\detJ=1$ with a zero energy change is not sufficient for BIBO stability; the required property is coercive-component membership. The blind gated channel and the state-replacing map also coincide exactly on all failures. This shows that the certificate screen, not the jump itself, provides the bound.

\subsection{Differentiating Access Mechanisms}
\label{sec:testbed}
To compare the mechanisms, we build a controlled reach battery on a small analytic testbed that does not depend on the architecture. The testbed has a $K$-basin (symmetric) double well along the reach coordinate, whose barriers span the causal box $L=2.5$ ($T=100$, $\varepsilon=0.05$, $c=1$, heavy reach coordinate $M_0=4.0$), and an $SO(2)$ sector for the latch. All results in this section use latent dimensions 2 and 4, 5 random seeds, designed channel placement, single-CPU execution, a sharp-boundary $\gamma=0$ rollout, and a mass band of $[4.0,0.25]$, which gives a $16\times$ inertial contrast. This controlled setting isolates the exactness of the theory; learning where to place entrances is outside the current scope. We set the governor so that plain relaxation cannot escape (kinetic energy $0.72 < \Delta V_b=1$). We measure landing rates at distances $d<L$, $d\in\{0.8,1.6,2.4\}$, and $d>L$, $d\in\{3.2,4.0,5.0\}$ (Table~\ref{tab:reach}; per-distance grid in App.~\ref{app:reach}).

\begin{table}[htb!]\centering\footnotesize
\caption{Reach battery summary: landing rates inside ($d{<}L$) and beyond ($d{>}L$) the causal box, measured $\detJ$, and diagnostic. The $(1{-}\gamma)^d$ entries hold for fixed $\gamma$.}
\label{tab:reach}
\begin{tabular}{lccll}
\toprule
Algorithm & $d{<}L$ & $d{>}L$ & Measured $\detJ$ & Diagnostic \\
\midrule
Plain relaxation & $0$ & $0$ & $(1{-}\gamma)^d$ & Escape-blocked \\
Squeeze $S^{(M)}$ & $1$ & $\to0$ & $1.000\pm4\mathrm{e}{-6}$ & Reach beyond $C_T$ needs energy $\le e^{2|\zeta|}H$ \\
Gated channel & $1$ & $1$ & $1.0$ exact & Flat access scaling; energy change $0.0$ \\
Newtonian control & $1$ & $1$ & $1.0$ & Confirms relativistic cap bounds flow \\
State-replacing map & $1$ & $1$ & $0.0$ exact & State overwritten; volume annihilated \\
Dense/$V$-throat & $\to0$ & $0$ & $(1{-}\gamma)^d$ & Modulates barrier; fails global reach \\
\bottomrule
\end{tabular}
\end{table}

\begin{figure}[htb!]
\centering
\includegraphics[width=\textwidth]{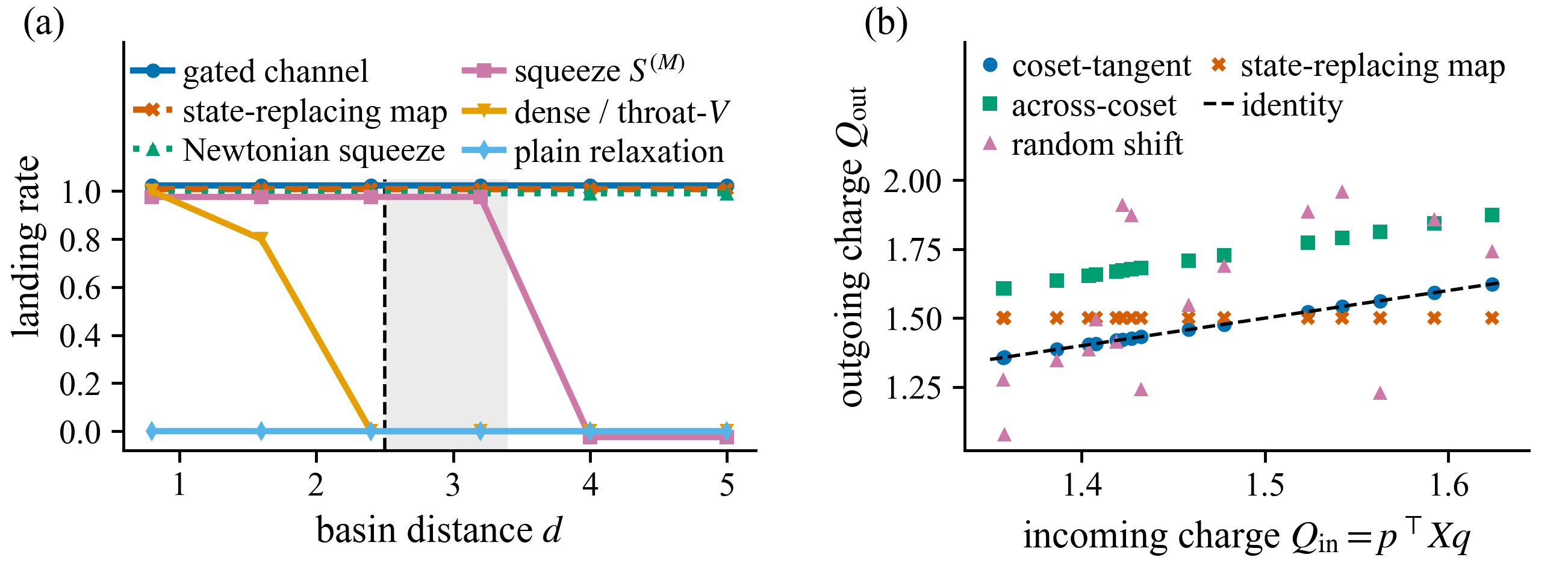}
\caption{Verification of the certificate stack on the designed testbed (Sec.~\ref{sec:testbed}). (a) Reach: landing rate against basin distance $d$ on a symmetric double well (dim 2, 5 seeds, $L=2.5$). (b) The certificate in force: outgoing against incoming Goldstone charge for 16 states drawn from the capture ball. Dashed line in (a): causal box $L$; grey band: crossover bracket. In (b), coset-tangent and across-coset are gated-channel jumps; the dashed line is $Q_{\rm out}=Q_{\rm in}$.}
\label{fig:reach}
\end{figure}

Squeeze reach follows from the launch state of the reach battery ($q=0$, momentum $p_0$ along the heavy coordinate), where the squeeze produces the instantaneous displacement $q_0'=(p_0/M_0)\sinh\zeta$, and the subsequent governed flow advances by at most the box half-width $L$. The squeeze therefore extends the reach bound to $[L,\,L+p_0\sinh\zeta/M_0]$ (shaded in Fig.~\ref{fig:reach}(a)). Landing at distance $d$ within the tolerance $0.4$ requires $\zeta\ge\operatorname{arcsinh}\bigl((d-0.4-L)\,M_0/p_0\bigr)$ (App.~\ref{app:deriv:squeeze}). At $p_0=1.2$ this gives $\zeta\ge2.0105$ for $d=4.0$ and $\zeta\ge2.64$ for $d=5.0$. Both exceed the swept budget $\zeta\le2.0$, so the squeeze row drops to zero at $d\ge4.0$ while $d\le3.2$ remains covered. The squeeze misses the outer distances because they need more than the swept rapidity budget, not because it cannot reach them with enough energy. Its energy grows as $e^{2|\zeta|}H$, exponentially in rapidity $\zeta$ and quadratically in the excess distance; the energy is \emph{not} exponential in the distance itself (App.~\ref{app:deriv:squeeze}). We validate the squeeze at several rapidities, $(\zeta,\text{ratio},\text{bound})=(0.25,1.13,1.65), (0.5,1.55,2.72), (1.0,3.79,7.39), (2.0,27.5,54.6)$, with $\det S^{(M)}=1.000\ (\pm4\times10^{-6})$ throughout. The $e^{2|\zeta|}$ bound is a matched-quadratic certificate; on a quartic well the raw ratio can exceed it.

We test the reach of the gated channel on the symmetric double well, with channel $-a\to+a$. The energy change is exactly $0.0$, and the channel translates the state to $q=+1.000$ with $\detJ=1.000$. The gated channel's energy cost is the endpoint potential difference, which is zero on this symmetric testbed, so here it does not grow with distance (Fig.~\ref{fig:reach}(a)). The untrained state-replacing map $(q,p)\mapsto(b,p)$ lands as often as the gated channel. However, its Jacobian is block-diagonal with zero blocks, so $\detJ=0$ exactly (forward-mode automatic differentiation): the map annihilates volume, is not invertible, and erases the state. It is an untrained analytic constant map, distinct from the learned decision head of \S\ref{sec:routing}. The Newtonian control reaches past the box, which confirms that the relativistic cap limits the flow. Plain relaxation is escape-blocked at all distances.

\subsubsection{The Certificate's Consequence: State Erasure vs.\ Transport}
\label{sec:erasure}
To measure the consequence of the $\detJ$ certificates, we track 16 states drawn uniformly from the capture ball ($\mathrm{std}(Q_{\rm in})=0.0803$) and compare the channel with an untrained state-replacing map that overwrites the state (Table~\ref{tab:latch}).

The channel's transport guarantee holds within a branch (gate fixed), where its canonical translation is injective. It shifts every incoming charge by the constant $p^\top X\Delta$ (max error $\le1.2\times10^{-7}$) and preserves the spread ($\mathrm{std}(Q_{\rm out})=0.0803$, Fig.~\ref{fig:reach}(b)). The pre-jump state is recoverable as $q_{\rm in}=q_{\rm out}-\Delta$ (max error $2.2{\times}10^{-8}$). We do not show that the full routing policy, which chooses between branches, is invertible. We observe a zero shift $\Delta Q=0.0$ (predicted precision $7.5\times10^{-10}$) for a coset-tangent channel and an across-coset shift of $\Delta Q=0.2500=p^\top X\Delta$. The matched channel, a coset-tangent $\Delta$ that also meets the energy constraint, guarantees stable transit.

\begin{table}[htb!]\centering\footnotesize
\caption{Latch transport on 16 states from the capture ball.}
\label{tab:latch}
\begin{tabular}{lcccccc}
\toprule
Mechanism & $\detJ$ & $\Delta Q$ mean & $\Delta Q$ std & Predicted $p^\top X\Delta$ & Max Error & $\mathrm{std}(Q_{\rm out})$ \\
\midrule
Gated channel (coset-tangent) & $1.0000$ & $-0.0000$ & $0.0000$ & $0.0000$ & $1.2\mathrm{e}{-7}$ & $0.0803$ \\
Gated channel (across-coset) & $1.0000$ & $0.2500$ & $0.0000$ & $0.2500$ & $0.0$ & $0.0803$ \\
Random shift & $1.0000$ & $0.0972$ & $0.2465$ & \emph{N/A} & --- & $0.2793$ \\
State-replacing map & $0.0000$ & $0.0379$ & $0.0803$ & \emph{N/A} & --- & $0.0$ \\
\bottomrule
\end{tabular}
\end{table}

On this designed testbed, the $\detJ=0$ state-replacing map instead sends all 16 states to one bit-identical charge ($\mathrm{std}(Q_{\rm out})=0$) and erases the coset content irreversibly. Volume preservation alone is not the latch certificate: a random shift also has $\detJ=1$ but scrambles the spread and erases the latch, whereas the matched channel preserves it.

\section{Translating Mechanisms to Learned Memories}
\label{sec:learned}
The testbed potentials are designed, but a learned memory's potential is not. We therefore test the mechanisms on trained associative memories and report where baselines outperform them. We first calibrate a compute-rationing gate, then map it across capacity and training budget, and finally compare non-local routing with baselines.

\subsection{Calibrated Compute-Rationing}
\label{sec:rationing}
We implement a self-calibrating gate on a trained CLU associative memory for MQAR key-value recall (vocab 256, capacity kv$\in\{16,24,32\}$, 5 seeds). At write time the memory probes itself with jittered copies of its stored keys and fits a per-model head that maps the relaxation residual to an error probability. A Learn-then-Test (LTT; \citealp{angelopoulos_learn_2025}) wrapper sets exit thresholds on a relaxation ladder \citep{geifman_selective_2017}. The wrapper held on all $30$ cells, with coverage $0.647\pm0.063$ at target risk $\varepsilon=0.05$. This rests on an exchangeability assumption between write-time probes and deployment queries, which our jittered protocol only approximates. The residual is a valid gate feature but adds at most $+0.024$ AUROC over a standard readout margin, and is therefore not the right exit threshold.

Rationing helps most at low capacity: at an under-converged 400-epoch budget, the gate reduces relaxation steps by $4.81\pm0.44\times$ at kv16 relative to an always-full budget, with higher accuracy. The gain shrinks as capacity grows: $1.57\pm0.07\times$ at kv24 and $1.14\pm0.06\times$ at kv32. At full convergence (2000 epochs) the gate matches the full-budget baseline rather than exceeding it. The gate's value is therefore compute rationing on clean retrieval, not general accuracy gains.

\subsection{Capacity and Training-Budget Map}
\label{sec:regime}
\begin{figure}[htb!]\centering
\includegraphics[width=0.82\linewidth]{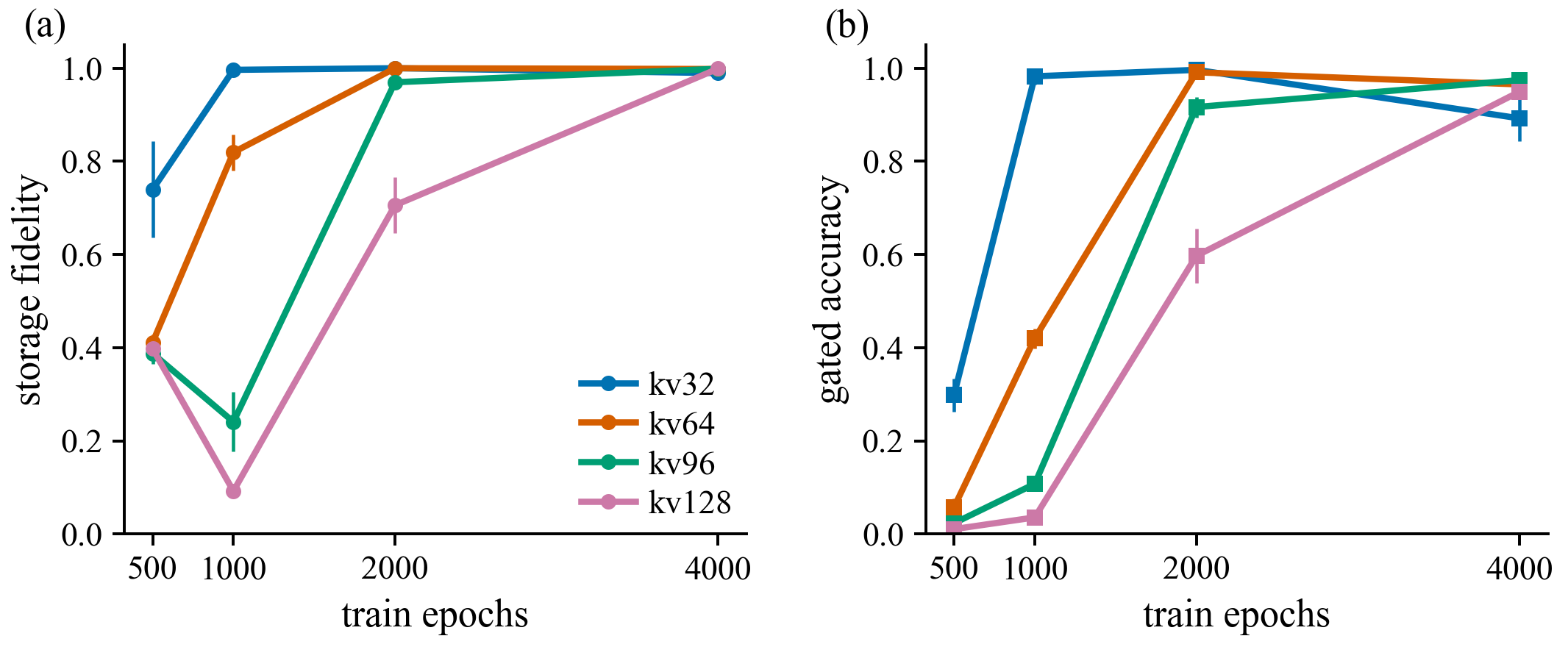}
\caption{Epoch-scaling frontier: CLU-EBM
storage fidelity and gated accuracy against training budget for kv$\in\{32,64,96,128\}$.
Required epochs scale with capacity rather than saturating at a fixed ceiling, and kv$32$
over-trains between $2000$ and $4000$ epochs. Mean over 3 seeds (42--44).}
\label{fig:frontier}
\end{figure}

We map the gate across capacity and training budget on a 198-job grid (Fig.~\ref{fig:frontier}, Table~\ref{tab:regime}). Storage fidelity converges to near-perfect levels with training, so the initial underperformance was an under-training artifact. Beyond kv64 the epoch budget, not a hard capacity limit, constrains performance; at 4000 epochs smaller cells over-train, and accuracy falls from $1.00$ to $0.89$. The required number of epochs grows with capacity: kv32 saturates by 1000 epochs, while kv96 needs 4000 epochs to reach an accuracy of $0.975$. The gate's step reductions ($9.9\times, 9.5\times, 6.2\times$ at kv32, kv64, kv96) are intra-CLU rationing against a full-budget CLU baseline. Under evaluation cue noise ($\sigma\in\{0.3,0.6\}$) the gate degrades sharply: at $\sigma=0.6$ on kv32 its accuracy is $0.36$, while CLU storage fidelity stays at $1.0$. The governed relaxation over-commits to the corrupted cue, and we record this noise wall as an open problem.

\begin{table}[htb!]\centering\footnotesize
\caption{Gate across capacity and training budget.}
\label{tab:regime}
\begin{tabular}{lccccc}
\toprule
Cell (N,kv) & Epochs & CLU Fidelity & Gate Acc & Intra-CLU Step-Reduction \\
\midrule
N128/kv32 & 500 & $0.76\pm0.09$ & $0.31\pm0.04$ & $1.2\times$ \\
N128/kv32 & 2000 & $1.00\pm0.00$ & $1.00\pm0.00$ & $9.9\times$ \\
N256/kv64 & 500 & $0.43\pm0.05$ & $0.06\pm0.02$ & $1.0\times$ \\
N256/kv64 & 2000 & $1.00\pm0.00$ & $0.99\pm0.00$ & $9.5\times$ \\
N384/kv96 & 500 & $0.40\pm0.03$ & $0.02\pm0.01$ & $1.0\times$ \\
N384/kv96 & 2000 & $0.97\pm0.01$ & $0.91\pm0.02$ & $6.2\times$ \\
\bottomrule
\end{tabular}
\end{table}

\subsection{Non-Local Routing}
\label{sec:routing}
On a lattice routing task (Table~\ref{tab:routing}), a one-hop non-local edge keeps a flat FLOP count as distance grows, unlike multi-hop diffusion. However, a parameter-matched, physics-free router outperforms the energy-gated edge in both FLOPs and accuracy. A learned decision router may select whether to use the edge, but the data still move through the certified $\detJ=1$ channel.

\begin{table}[htb!]\centering\footnotesize
\caption{Routing: local and distant accuracy (L/D) and floating-point operations per query over 5 seeds.}
\label{tab:routing}
\begin{tabular}{@{}lcccc@{}}
\toprule
Algorithm & $N{=}4$ Acc (L/D) & $N{=}4$ FLOP & $N{=}8$ Acc (L/D) & $N{=}8$ FLOP \\
\midrule
Local only & $0.500$ ($1.00$/$0.00$) & $5.88\mathrm{e}7$ & $0.500$ ($1.00$/$0.00$) & $5.88\mathrm{e}7$ \\
Gated energy & $0.887\pm0.139$ ($0.90$/$0.88$) & $1.18\mathrm{e}8$ & $0.715\pm0.172$ ($0.82$/$0.61$) & $1.18\mathrm{e}8$ \\
Dense network & $0.500$ ($0.00$/$1.00$) & $1.18\mathrm{e}8$ & $0.448\pm0.070$ ($0.00$/$0.90$) & $1.18\mathrm{e}8$ \\
Diffusion chain & $0.652\pm0.159$ ($0.90$/$0.41$) & $1.76\mathrm{e}8$ & $0.548\pm0.138$ ($0.82$/$0.28$) & $2.94\mathrm{e}8$ \\
Physics-free router & $1.000\pm0.000$ ($1.00$/$1.00$) & $8.81\mathrm{e}7$ & $0.948\pm0.070$ ($1.00$/$0.90$) & $8.81\mathrm{e}7$ \\
\bottomrule
\end{tabular}
\end{table}

\section{Discussion}
For a conservative memory built as an LDM, test-time compute is certified access: under the stated assumptions, the dynamics fix the volume, energy and latch consequences of every access mechanism. A squeeze resolves escape with bounded injection, at a cost exponential in rapidity and quadratic in the excess distance. A gated channel reaches globally at $\detJ=1$ exactly, at an energy cost equal to the endpoint potential difference (zero on the matched channel here, so independent of distance in this testbed), with exact latch transport at the jump. These guarantees are conditional: the volume certificate assumes damping frozen within a step and does not cover the state-dependent governor, the latch law holds at the jump only, and the channel is injective within a branch. They also do not bound trajectories: BIBO stability requires coercive-component screening (\S\ref{sec:bibo}). On learned memories, the mechanisms ration compute on clean retrieval, while a physics-free router is stronger for non-local routing. The main contribution is the theoretical formalization of the certificate stack and the empirical mapping of its limits; we do not claim a general leaderboard advantage. App.~\ref{app:negative} records the main negative results.

With a quadratic payload channel and a fixed read budget, this implementation does not achieve exact value fidelity, amplitude-independent address hold and amplitude-independent read latency at once; we observe this as a design trade-off of the implementation, not a proved incompatibility. The implementation gives up one of these, so a faded memory requires more integration steps to read. Future work includes learned entrance-steering for the gated channels, retry acceptance as a certified kernel (App.~\ref{app:mcmc}), and tests on learned latent potentials at scale.

\bibliographystyle{plainnat}
\bibliography{refs}

\appendix

\section{Flag-Provenance Tables}
\label{app:flags}
All computational results use the non-default configurations listed below. We ran all computational experiments on standard laptop CPUs.

\subsection{Certificate Stack Configuration}
\noindent{\footnotesize
\begin{tabular}{@{}p{0.24\textwidth}p{0.70\textwidth}@{}}
\toprule
Flag & Value \\
\midrule
JAX & 0.9.0 \\
Kinetic modes & relativistic (reach, gated channel, throat); \texttt{newtonian\_learned} (Newtonian control) \\
Mass band & $[4.0, 0.25]$ yielding $M_{\rm eff,0}=4.0$, $16\times$ contrast \\
$c$ / rest mass / $m_0$ & 1.0 / 1.0 / 1.0 \\
Time step dt / Horizon $T$ & 0.05 / 100 yielding box $L=T\varepsilon c/\sqrt{M_0}=2.5$, $v_{\max,0}=0.5$ \\
Damping $\gamma$ & reach rollout $\gamma=0$ (sharp box) \\
Rapidity $\zeta$ grid & $[0,0.1,0.2,0.3,0.4,0.6,0.8,1.0,1.5,2.0]$ \\
Basin geometry & double well along coord 0, barrier $\Delta V_b=1.0$, $d\in\{0.8,1.6,2.4,3.2,4.0,5.0\}$ \\
Initialization & $q_0=0$, $p_0=(1.2,0,\dots)+0.02\,\mathcal N$ (kinetic energy below barrier) \\
Landing criterion & $\min_{\rm traj}|q_0-d|<0.4$ \\
Seeds & reach $\{0,1,2,3,4\}$; latch \texttt{default\_rng(0)} \\
Dimensions & 2 and 4 \\
\bottomrule
\end{tabular}}

\subsection{Calibrated Gate Configuration}
\noindent{\footnotesize
\begin{tabular}{@{}p{0.24\textwidth}p{0.70\textwidth}@{}}
\toprule
Flag & Value \\
\midrule
Task & MQAR CLU-EBM, vocab 256, capacity kv$\in\{16,24,32\}$ \\
Seeds & 5 (base 42); 90 models \\
Write configuration & kinetic relativistic, potential mlp, hidden 128, epochs 400, lr $1\mathrm{e}{-3}$, batch 16, k\_steps 50, buffer 128, friction 0.3, temperature 0.3, input noise $\sigma$ 0.05 \\
Gate configuration & calib features \texttt{r\_margin}, learned $\tau$ at write time, $p_{\rm exit}=0.5$; cost checkpoints $300/1200/2100/3000$ \\
Learn-then-Test & fixed-sequence, exact binomial p-values, targets $\varepsilon\in\{0.05,0.1\}$ \\
Hopfield baseline & Platt-calibrated logit margin, identically probed \\
\bottomrule
\end{tabular}}

\subsection{Routing Configuration}
\noindent{\footnotesize
\begin{tabular}{@{}p{0.24\textwidth}p{0.70\textwidth}@{}}
\toprule
Flag & Value \\
\midrule
Lattice and task & distance $N\in\{4,8\}$, embed\_dim 12, vocab 128 \\
Write configuration & kinetic relativistic, potential mlp (coercive), hidden 128, epochs 400, lr $1\mathrm{e}{-3}$ \\
Retrieval routing & dt 0.05, relax\_steps 250, governor\_sensitivity 0.95 \\
Router MLP & hidden 32 (449 params), epochs 300, lr $3\mathrm{e}{-3}$ \\
Workload mixes & $[[0.5,0.5],[0.8,0.2],[0.95,0.05]]$ \\
\bottomrule
\end{tabular}}

\subsection{Regime Map Configuration}
\noindent{\footnotesize
\begin{tabular}{@{}p{0.24\textwidth}p{0.70\textwidth}@{}}
\toprule
Flag & Value \\
\midrule
Training epochs & swept across $\{500,1000,2000,4000\}$ \\
Kinetic and potential & relativistic / mlp (coercive) \\
Architecture & embed\_dim 16 (CLU dim 32) / hidden 128 / vocab 256 \\
Retrieval & relax\_steps 300, governor\_sensitivity 0.95, calib \texttt{r\_margin} \\
Seeds & $\{42,43,44,45,46\}$ \\
\bottomrule
\end{tabular}}

\subsection{Latch Transport and BIBO Battery Configuration}
\noindent{\footnotesize
\begin{tabular}{@{}p{0.24\textwidth}p{0.70\textwidth}@{}}
\toprule
Flag & Value \\
\midrule
Training & none; analytic potentials, oracle channel placement \\
Kinetic mode & relativistic (reach, gated channel, throat, BIBO) \\
Damping $\gamma$ & reach 0 (sharp box); BIBO 0.02 (bounded arms must settle) \\
BIBO potential & $k=1.0$, $\epsilon=0.02$ yielding $x_b=3.5355$, barrier $V_b=3.125$ \\
BIBO parameters & exits $b\in\{1.0,2.0,3.0,3.6,4.0,5.0\}$; horizon $T=2000$ \\
Latch transport & 16 incoming states, ball radius $0.3<\rho=0.35$ \\
Seeds & reach/BIBO $\{0,1,2,3,4\}$; latch cloud \texttt{default\_rng(0)} \\
\bottomrule
\end{tabular}}

Two configuration axes most affect comparative performance: damping and training epochs. The calibrated-gate evaluation uses $\gamma=0.3$ and 400 epochs, an under-converged regime. The regime map instead sweeps up to 4000 epochs to isolate converged behavior.

\section{Reach Battery Full Grid}
\label{app:reach}
We use 5 seeds and causal box $L=2.5$.
\begin{center}\footnotesize
\begin{tabular}{lcccccc}
\toprule
Algorithm & $d{=}0.8$ & $1.6$ & $2.4$ & $3.2$ & $4.0$ & $5.0$ \\
\midrule
Plain relaxation & 0 & 0 & 0 & 0 & 0 & 0 \\
Squeeze $S^{(M)}$ & 1 & 1 & 1 & 1 & 0 & 0 \\
Gated channel & 1 & 1 & 1 & 1 & 1 & 1 \\
Newtonian control & 1 & 1 & 1 & 1 & 1 & 1 \\
State-replacing map & 1 & 1 & 1 & 1 & 1 & 1 \\
Dense/$V$-throat & 1 & 0.8 & 0 & 0 & 0 & 0 \\
\bottomrule
\end{tabular}
\end{center}

\section{Documented Negative Results}
\label{app:negative}
We record the main negative findings to mark the limits of our mechanisms. First, squeeze retries recover no performance at single-unit scale when selecting among already reachable attractors. The squeeze therefore addresses kinematic reach failure, not local selection. Second, the compute dial does not give robustness to corrupted cues. Gate accuracy degrades sharply under evaluation cue noise while storage fidelity stays at $1.0$, so the governed relaxation over-commits to the corrupted cue rather than failing to store the pattern. Noise-aware thresholds, longer relaxation budgets and denoising initialization are open work. Finally, the raw energy signal adds negligible predictive value over a standard readout margin. Our framework therefore relies on the calibrated rationing mechanism itself, not on a superior physics signal.

\section{Analytic Verification Protocols}
We ran all verifications in Python at standard numerical precision on CPU. The causal cap of the reach box matches its prediction to a relative error of $2.0\times10^{-12}$. The Jacobian determinant matches $1.000$, and its state-dependent deviations match the analytical prediction to $2.2\times10^{-16}$. The Goldstone latch transit error stays below $1.8\times10^{-15}$. The governor re-absorption matches the predicted exponential injections at all swept rapidities.

\section{Certified Markov Kernels: Derivations and Design Rules}
\label{app:mcmc}
To obtain a stationarity certificate for test-time retries, we derive four operating constraints, which are also the design rules of the MCMC implementation. First, the symmetric Lorentz squeeze is exactly volume-preserving ($\detJ=1$), so it needs no Jacobian correction in the acceptance ratio. A retry that proposes a sign-symmetrized squeeze and accepts with $\min(1,e^{-\Delta H/T})$ forms a detailed-balance kernel for $e^{-H/T}$ \citep{duane_hybrid_1987}. The state-dependent governor, however, is non-volume-preserving dissipation. The deployed sequence is therefore Metropolis-within-annealing: it converges to a colder maximum-a-posteriori state and does not act as a Gibbs sampler, and we do not claim stationarity for the governed composite.

Second, the orbit of the squeeze family is a single hyperbola, so the squeeze alone is reducible and not ergodic. Ergodicity requires momentum refreshment: the squeeze must be layered with a Metropolis-Adjusted Langevin Algorithm (MALA) step \citep{roberts_exponential_1996} to mix across the state space \citep{brooks_mcmc_2011}.

Third, the Langevin noise scale must be calibrated with the discrete fluctuation-dissipation relation, $\sigma_i^\star=\sqrt{M_{{\rm eff},i}T\gamma(2-\gamma)}$, to prevent finite-budget shadow bias. This calibration reduces the finite-budget shadow-bias error from $0.0995$ to $0.0065$. In relativistic kinetic modes the target distribution follows Maxwell-J\"uttner statistics. Fixed-covariance Gaussian refreshes are therefore invalid, and only the Metropolis-Hastings corrections maintain stationarity.

Finally, the Gibbs measure is flat along Goldstone cosets, so any coset-tangent proposal has zero energy difference and is always accepted. These proposals generate an unbiased random walk that diffuses and erodes the stored register at a rate proportional to the proposal-scale variance. Projecting proposals orthogonal to the coset tangent suppresses this drift, and is required to prevent this topological degradation. This shows that without projection, even a certified MCMC retry progressively erases stored memory contents.

\section{Derivation of the Bounded-Injection Certificate}
\label{app:deriv:squeeze}
This appendix gives the full derivation of the squeeze certificate $H(S_\zeta z)\le e^{2|\zeta|}H(z)$ and of the reach bracket used in \S\ref{sec:testbed}; the causal box and the hard-gate Jacobian are derived in \S\ref{sec:setup} and \S\ref{sec:mech}. The causal box uses the relativistic kinetic energy, while the squeeze energy bound is proved for a Hamiltonian quadratic in position and momentum (the matched quadratic). The two results therefore rest on different kinetic energies and are not a single joint statement.

To define the squeeze map on the targeted coordinate pair $(q_i,p_i)$, we pass to mass-weighted coordinates $u=\sqrt{M_i}\,q_i$, $v=p_i/\sqrt{M_i}$ and let the mass-weighted squeeze $S^{(M)}_\zeta$ act as the hyperbolic rotation
\[
u'=u\cosh\zeta+v\sinh\zeta,\qquad v'=v\cosh\zeta+u\sinh\zeta,
\]
and as the identity on all other coordinates. In the original variables, $q_i'=q_i\cosh\zeta+(p_i/M_i)\sinh\zeta$ and $p_i'=p_i\cosh\zeta+M_iq_i\sinh\zeta$. The map is linear with determinant $\cosh^2\zeta-\sinh^2\zeta=1$ on the active pair and $1$ elsewhere, hence exactly symplectic. This is the $\det S^{(M)}=1$ certificate.

To derive the displacement law and the reach bracket, we apply the squeeze at the launch state of the reach battery ($q=0$, momentum $p_0$ along the heavy coordinate). It produces the instantaneous displacement $q_0'=(p_0/M_0)\sinh\zeta$, and the subsequent governed flow advances by at most the box half-width $L$ (\S\ref{sec:setup}). The squeeze-extended reach is therefore contained in the bracket $[\,L,\;L+p_0\sinh\zeta/M_0\,]$ of Figure~\ref{fig:reach}. Landing at distance $d$ within the tolerance $0.4$ of the flag table requires $(p_0/M_0)\sinh\zeta\ge d-0.4-L$, i.e.
\[
\zeta\ \ge\ \operatorname{arcsinh}\!\bigl((d-0.4-L)\,M_0/p_0\bigr),
\]
which at $L=2.5$, $p_0=1.2$, $M_0=4.0$ evaluates to $\zeta\ge2.0105$ for $d=4.0$ and $\zeta\ge2.64$ for $d=5.0$. Both exceed the swept budget $\zeta\le2.0$. This is why the squeeze row of the reach battery drops to zero at $d\ge4.0$ while $d\le3.2$ remains covered.

The energy-bound certificate is a matched-quadratic statement: on the active pair the quadratic Hamiltonian is isotropic in the same mass-weighted coordinates, $H_{\mathrm q}=\half\kappa\,(u^2+v^2)$ with $\kappa>0$. Under the squeeze,
\[
u'^2+v'^2=(u^2+v^2)\cosh2\zeta+2uv\,\sinh2\zeta
\ \le\ (u^2+v^2)\bigl(\cosh2\zeta+|\sinh2\zeta|\bigr)
=(u^2+v^2)\,e^{2|\zeta|},
\]
using $2|uv|\le u^2+v^2$. Hence $H(S_\zeta z)\le e^{2|\zeta|}H(z)$, with equality exactly on the diagonal $u=\pm v$ (sign matched to that of $\zeta$). The injected energy obeys $\Delta H\le(e^{2|\zeta|}-1)H$, the form quoted in the certificate table. Spectator coordinates only tighten the bound: they contribute the same non-negative energy to both sides and pull the ratio toward one. The text uses two consequences. First, at the launch state $u=0$ the ratio equals $\cosh2\zeta$ exactly, so the measured ratios must lie between $\cosh2\zeta$ and the bound $e^{2|\zeta|}$, as they do. Second, the bound is attained only on the diagonal, so generic states sit strictly below it.

We next show why the energy is exponential in rapidity and quadratic in the excess distance. Writing $\delta=(p_0/M_0)\sinh\zeta$ for the displacement beyond $L$, the launch-state ratio is
\[
\cosh2\zeta=1+2\sinh^2\zeta=1+2\,(M_0\delta/p_0)^2,
\]
exactly quadratic in $\delta$. The certificate $e^{2|\zeta|}=(\cosh\zeta+|\sinh\zeta|)^2$ is exponential in the rapidity and also quadratic in $\delta$ at large $\zeta$. This is the scaling stated in the abstract; the energy is \emph{not} exponential in the distance itself.

The scope of the inequality is the matched quadratic $H$, for which it is proved. On a quartic well the potential term transforms with higher powers of $\cosh\zeta$, so the raw ratio can exceed $e^{2|\zeta|}$, as reported.

\end{document}